\documentclass[letterpaper]{article} % DO NOT CHANGE THIS
\usepackage{aaai25}  % DO NOT CHANGE THIS
\usepackage{times}  % DO NOT CHANGE THIS
\usepackage{helvet}  % DO NOT CHANGE THIS
\usepackage{courier}  % DO NOT CHANGE THIS
\usepackage[hyphens]{url}  % DO NOT CHANGE THIS
\usepackage{graphicx} % DO NOT CHANGE THIS
\usepackage{natbib}  % DO NOT CHANGE THIS AND DO NOT ADD ANY OPTIONS TO IT
\usepackage{caption} % DO NOT CHANGE THIS AND DO NOT ADD ANY OPTIONS TO IT
\usepackage{algorithm}
\usepackage{algorithmic}
\usepackage{enumitem}
\usepackage{newfloat}
\usepackage{listings}
\usepackage{tabulary}
\usepackage{amsmath}
\usepackage{booktabs}
\usepackage{multirow}
\usepackage{float}
\usepackage{nameref}
\usepackage{indentfirst}
\usepackage[capitalize]{cleveref}
\usepackage{xcolor}
\usepackage{tcolorbox} % for the rounded box
\tcbuselibrary{skins} % to use skin options like rounded corners
\definecolor{LightBlue}{rgb}{0.68, 0.85, 0.9} % This is an approximation of light blue

\usepackage{newfloat}
\usepackage{listings}
\DeclareCaptionStyle{ruled}{labelfont=normalfont,labelsep=colon,strut=off} % DO NOT CHANGE THIS
\floatstyle{ruled}
\newfloat{listing}{tb}{lst}{}
\floatname{listing}{Listing}
\title{Adapting to Non-Stationary Environments: Multi-Armed Bandit Enhanced Retrieval-Augmented Generation on Knowledge Graphs}
\author{
    Xiaqiang Tang\textsuperscript{\rm 1,2},
    Jian Li\textsuperscript{\rm 2}\thanks{Corresponding author},\\
    Nan Du\textsuperscript{\rm 2},
    Sihong Xie\textsuperscript{\rm 1,*},
}
\affiliations{
    \textsuperscript{\rm 1}The Hong Kong University of Science and Technology (Guangzhou)\\
    \textsuperscript{\rm 2}Tencent Hunyuan\\
    
}

\usepackage{bibentry}
\begin{document}

\maketitle

\begin{abstract}
Despite the superior performance of Large language models on many NLP tasks, they still face significant limitations in memorizing extensive world knowledge. 
Recent studies have demonstrated that leveraging the Retrieval-Augmented Generation (RAG) framework, combined with Knowledge Graphs that encapsulate extensive factual data in a structured format, robustly enhances the reasoning capabilities of LLMs.
However, deploying such systems in real-world scenarios presents challenges: the continuous evolution of non-stationary environments may lead to performance degradation and user satisfaction requires a careful balance of performance and responsiveness.
To address these challenges, we introduce a Multi-objective Multi-Armed Bandit enhanced RAG framework, supported by multiple retrieval methods with diverse capabilities under rich and evolving retrieval contexts in practice.
Within this framework, each retrieval method is treated as a distinct ``arm''. The system utilizes real-time user feedback to adapt to dynamic environments,  by selecting the appropriate retrieval method based on input queries and the historical multi-objective performance of each arm.
Extensive experiments conducted on two benchmark KGQA datasets demonstrate that our method significantly outperforms baseline methods in non-stationary settings while achieving state-of-the-art performance in stationary environments. Code and data are available at https://github.com/FUTUREEEEEE/Dynamic-RAG
\end{abstract}


\section{Introduction}

% Large language models (LLMs) \cite{chowdhery2023palm,openai2023gpt,touvron2023llama} have showcased remarkable performance across various natural language processing tasks \cite{bang2023multitask,brown2020language}. While these models excel in generating coherent and contextually appropriate responses, their performance on complex knowledge-intensive tasks remains unsatisfactory  \cite{petroni2020kilt,talmor2018web}. 
% LLMs may generate unfaithful information in conflict with factual knowledge suffering from hallucination \cite{hallucination,li2023halueval}, falling short of mastering domain-specific or real-time knowledge \cite{peng2023check}.
% To address these shortcomings, Retrieval-Augmented Generation (RAG) \cite{RAG} has been developed to improve LLM reasoning ability and has demonstrated notable progress in reducing hallucination effects and providing trustworthy responses with up-to-date knowledge \cref{RAG}. 

%RAG 必要性
Large language models (LLMs) \cite{chowdhery2023palm,openai2023gpt,touvron2023llama} excel in natural language processing tasks \cite{bang2023multitask,brown2020language} but struggle with knowledge-intensive challenges, often producing unfaithful or hallucinated information \cite{petroni2020kilt,hallucination}. Retrieval-Augmented Generation (RAG) \cite{RAG} has been developed to enhance LLM reasoning, effectively reducing hallucinations and providing reliable, up-to-date information. In this approach, when presented with a user query, a retriever first extracts relevant information from a knowledge base, which is then provided to the LLM to generate the final response.Recent advancements \cite{G-Retriever,Rog,thinkongraph,KAPING} in RAG systems have increasingly incorporated Knowledge graphs (KGs) \cite{baek2023knowledge,luo2023chatrule} as the underlying knowledge base. KGs store vast amounts of factual data in a structured format, which enables more dependable and systematic reasoning by LLMs.

% \begin{figure}[ht!]
%      \includegraphics[width=\linewidth]{fig/RAG-kg.png}
%      \caption{Comparative diagram of standalone Language Model (LLM Only) and Retrieval-Augmented Generation (LLM+RAG).}
%      \label{RAG}
% \end{figure}

% \begin{figure}[ht!]
%      \includegraphics[width=\linewidth]{fig/IR_methods_v2.png}
%      \caption{KG-based RAG systems present diverse retrieval methods.}
%      \vspace{-5mm}
%      \label{IR_methods}
% \end{figure}

\begin{figure*}[t]
     \includegraphics[width=\textwidth]{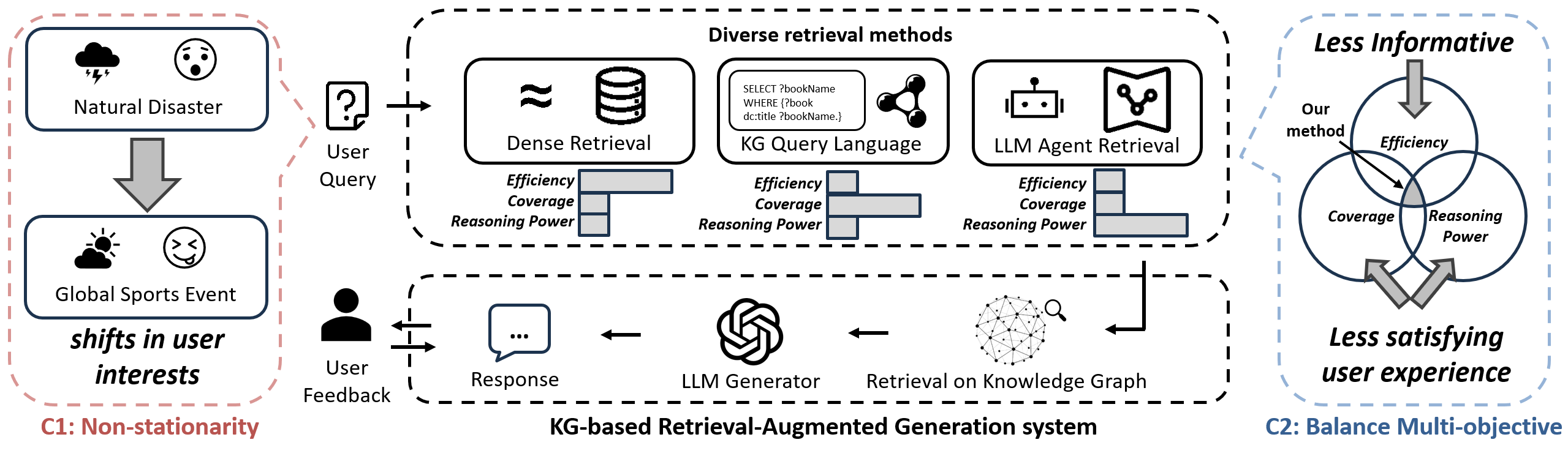}
     \caption{An online KG-based RAG system facing challenges from non-stationary environments and the need to balance multiple objectives for optimal user experience.}
     \vspace{-3mm}
     \label{background}
\end{figure*}

 Unlike unstructured text databases (e.g. Wikipedia), the organized nature of KGs provides diverse retrieval methods, with significantly different capabilities and costs. For example, dense retrieval methods \cite{bge,decaf} are typically fast but offer limited reasoning capabilities. In contrast, using LLMs to generate KG query languages (e.g., SPARQL) as in ChatKBQA \cite{chatkbqa} provides high coverage and is suitable for multi-entity retrieval. Methods like RoG \cite{Rog}, where LLMs function as search agents excel in complex reasoning. However, both methods require interactions with LLMs like ChatGPT \cite{ChatGPT}, leading to longer execution times. However, current KG-based RAG systems often rely solely on a single retrieval method or use static neural network routers \cite{NN-router,llamaIndexDefineSelector}, which require complete labeled data for supervision and periodic fine-tuning. Moreover, while RAG systems are often deployed in scenarios where users can provide feedback on generated responses \cite{gamage2024multi,alan2024rag}, current systems generally neglect this feedback. Relying on a single retrieval method, or computationally intensive ensemble all retrieval results can not ensure responses that are both timely and informative. On the other hand, static neural network routers cannot effectively leverage real-time feedback to continually adapt to changing user needs and system variability.

Therefore, deploying RAG systems in real-world scenarios faces the following challenges as described in \cref{background}:
\textcolor[HTML]{bb5c58}{\textbf{(C1)}}: Non-stationary environments require RAG systems to adapt to two sides continuously: on the user side, the evolving nature of queries driven by trending topics, and on the server side, the backend retrieval model upgrading.
\textcolor[HTML]{376092}{\textbf{(C2)}}: In practical applications RAG systems, such as personal home assistants and customer support chatbots, balancing multi-objective, such as efficiency, coverage, and reasoning power, is crucial to providing informative and satisfying user experiences. Failing to address the diverse demands of queries and deliver timely, comprehensive responses can result in less informative interactions or unsatisfying user experience.

%\textbf{用具体的词 manage-> aggregate}

In response to \textcolor[HTML]{bb5c58}{\textbf{(C1)}}, we proposed a RAG framework enhanced by deep contextual Multi-arm Bandit\cite{DeepMAB}, utilizing a lightweight language model as the backbone to interpret user queries and predict the suitability of each retrieval method. The model is updated on a per-query basis using feedback, ensuring robust performance and adaptability in non-stationary environments. % 不能用抽象的词，具体是如何解决,  Multi-objective,说具体一点
%多个agent是如何选择的
In response to \textcolor[HTML]{376092}{\textbf{(C2)}}, we incorporated the Generalized Gini Index to aggregate multi-objective user demands effectively, ensuring that no single objective dominates the other objective. By balancing retrieval coverage, accuracy, and response time, our framework enhances user experience by providing informative answers under time constraints.

Our main technical contributions are as follows:
\begin{itemize}
    \item We enhanced KG-based RAG systems by employing an MAB model for dynamic retrieval selection and continuous adaptation to non-stationarity using user feedback.

    \item We utilized the Generalized Gini Index to aggregate multi-objective rewards, ensuring both informative and timely responses.

    \item We evaluated our framework on two well-established KBQA datasets. Our results demonstrate that our methods significantly outperform baseline approaches in non-stationary environments and surpass state-of-the-art KG-based RAG systems in stationary settings.
\end{itemize}

\section{Method}

\begin{figure*}[t]
     \includegraphics[width=\textwidth]{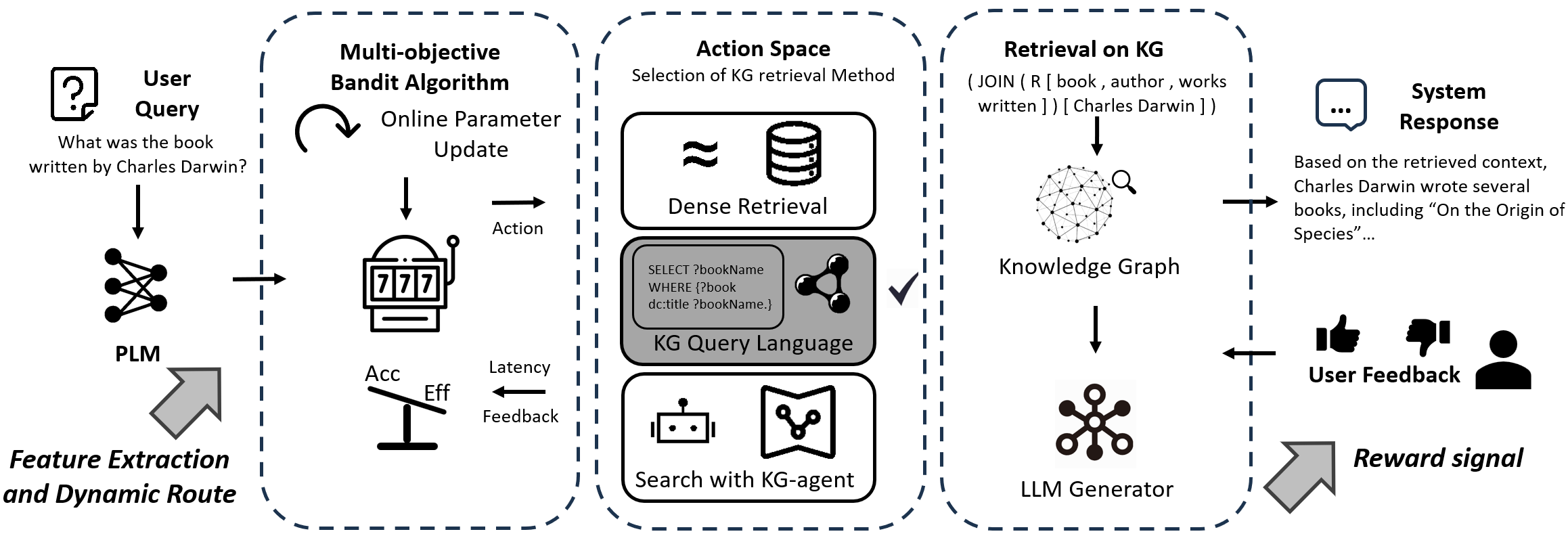}
     % \vspace{-4mm}
     \caption{Proposed MAB-enhanced RAG framework. The input query undergoes feature extraction (e.g., multi-entity query), followed by the MAB algorithm, which selects the optimal retrieval method by predicting the most rewarding option (e.g., Query Language method). The selected method retrieves information from a Knowledge Graph (KG), and an LLM generates the final response. Feedback is collected as a reward, updating the MAB model parameters online, and enabling continuous adaptation to non-stationary environments.}
     \vspace{-3mm}
     \label{method}
\end{figure*}

The diverse capabilities of different retrieval methods necessitate a strategic model for their selection.
Simply running multiple retrievers and then aggregating their results often proves sub-optimal due to two main factors: the need for timely responses and the disparate performance characteristics of various retrieval methods, as highlighted in \cref{tab:station_comp}. For example, while dense retrievers provide rapid responses, KG agent-based retrievers slow down the system due to LLM inference.

Consequently, we developed a model that dynamically assigns queries to the most suitable retrievers. Unlike static neural network routers, which require collecting complete labeled data for supervision (involving the execution of all retrieval methods) and periodic fine-tuning, limiting their adaptability to non-stationary environments. Our approach leverages real-time user feedback as a reward signal to update the model. This adaptability is crucial in the dynamic nature of RAG applications, such as shifting user interests and backend retriever upgrades requiring continuous optimization.

\subsection{Problem Setup}
The optimization of KG-based RAG systems employing multiple retriever backends and real-time feedback is structured as follows:

\begin{itemize}[leftmargin=*]
    \item Initially, the system receives a user input query context \(x\).
    \item The PLM model \(f_\theta\) processes the query and selects an action \(a\) from the action space \(A\), which includes \(K\) potential retrieval methods, each representing an arm in a multi-armed bandit.
    \item Upon selection, the system receives feedback on the performance of the chosen retrieval method \(a\) (e.g. 1 indicating a good response, 0 indicating a bad response), providing "partial-information" feedback. This limitation restricts the system’s ability to assess unselected methods.
    \item Utilizing this feedback, the model iteratively refines its strategy to improve base retrieval method selection for future queries.
\end{itemize}

\subsection{Deep Multi-objective Contextual Bandits}

\subsubsection{Query Encoding Model:}
In order to effectively select retrieval methods, it is crucial to discern patterns within user queries and associate these with the capabilities of suitable retrieval methods. Traditional linear models in contextual bandits \cite{linucb,musicMOMAB}, while effective in certain scenarios, often fall short due to the complex natural language patterns present in user queries.

To address limitations and ensure real-time service, we utilize the lightweight Pre-trained Language Model, DistilBERT \cite{distilbert}. As a streamlined version of BERT, DistilBERT retains approximately 97\% of BERT’s language understanding capabilities and increases processing speed by 60\%. This model provides a robust and nuanced approach to modeling natural language queries, which facilitates the precise identification of appropriate retrieval methods for individual queries and supports continuous real-time refinement through user feedback.

Specifically, our query encoding model, \( f_\theta \), uses DistilBERT to efficiently interpret natural language, taking a query as input context and producing an arm selection distribution \( z = f_\theta(x) \).

\subsubsection{Arm Selection Strategy:}
\label{ArmSelectionStrategy}
Upon receiving an action distribution estimation \( \mathbf{z} = f_\theta(x) \) from the encoding model, we employ an epsilon-greedy strategy \cite{epsilon-greedy} to balance the trade-off between exploration and exploitation \cite{UCB, auer2002finite}. This balance is crucial in ensuring that the system not only leverages the information gathered so far (exploit known retrieval methods that have proven effective) but also explores new possibilities to enhance learning (explore other retrieval methods that could potentially offer better results). Specifically:
\begin{itemize}[leftmargin=*]
    \item With a probability of \(1 - \epsilon\), the system selects the arm with the highest predicted reward, \(a = \max(\mathbf{z})\), based on the output from the encoding model.
    \item Conversely, with a probability of \(\epsilon\), the system explores by randomly selecting an arm, facilitating the discovery of potentially more effective retrieval methods.
\end{itemize}

This strategy enables the system to predominantly rely on the best-known actions to maximize immediate rewards while maintaining the flexibility to explore new possibilities. This approach is essential to mitigate the risk of converging to a locally optimal model due to partial information feedback, fostering the discovery of superior long-term solutions through randomized exploration.

\subsubsection{Learning Algorithm:}
% After selecting a retrieval method, our model updates based on the observation associated with the chosen method, but it does not have access to the information from methods not selected (i.e. partial information feedback). During training, we employ detailed evaluation results (including hit and recall metrics) to train the model, while in testing, we simulate an online environment using a binary hit value (0 or 1) to approximate user feedback.
After selecting a retrieval method, our model updates based on the observation associated with the chosen method, but it does not have access to the information from methods not selected (i.e., partial information feedback). Inspired by "offline-to-online" learning\cite{lee2022offline,guo2024sample}, we first pre-train the model in an offline environment to learn a robust initial strategy. Subsequently, we fine-tune the model in an online setting using partial user feedback, allowing it to adapt continuously to real-world conditions.

Traditional RAG systems often focus on optimizing model accuracy. However, real-world applications of RAG systems demand not only accuracy but also real-time responsiveness, introducing the need for multi-objective optimization. We use the Generalized Gini Index \cite{GGI} to balance system performance with retrieval time, ensuring both accuracy and efficiency are optimized simultaneously.

During training, we use detailed evaluation metrics, including informativeness measures like hit and recall, to optimize for accuracy and coverage. Retrieval latency is also used as feedback to enhance efficiency. In testing, we simulate an online environment with a hit value (0 or 1) to approximate binary user feedback. This offline-to-online learning approach ensures the model is well-prepared before deployment and can adapt effectively to dynamic user interactions.

\begin{algorithm}[t]
\caption{Deep GGI-MO bandit enhanced RAG learning algorithm}
\label{alg3}
\begin{algorithmic}[1]
\STATE \textbf{Input:} The query context set \(X\), pre-trained language model parameters \(\theta\).
\STATE \textbf{Initialize:} Set equal initial weights for the \(\textbf{w}\).

\FOR {\(x \in X\)}
    \STATE Encode the query \(x\) and obtain the estimated action distribution \( z = f_{\theta}(x) \).
    \STATE Select an retriever (arm) \(a\) based on the selection strategy described in \cref{ArmSelectionStrategy}.
    \STATE Observe the retrieval context to compute the loss components as per \cref{loss-components} and the execution time \(d\) for the retrieval process.
    \STATE Update the model weights \(\theta\) by minimizing the loss \( Loss_{GGI}(\theta) \) using gradient descent.
\ENDFOR
\STATE \textbf{Output:} Updated model weights \(\theta\).
\end{algorithmic}
\end{algorithm}

% \begin{equation}
%     l_{a}^1 = MSE(max(z), \hat{h})
%     l_{a}^2 = MSE(max(z), \hat{rc}
%     l_{a}^3 = KLDiv(z, \sigma(d_i))
% \end{equation}

% The methodology of our approach is encapsulated in Algorithm \ref{alg3}. In our approach, we focus on three distinct optimization objectives to comprehensively evaluate the performance of the retrieval methods. Towards improving the overall performance of the whole RAG system, for each query, we evaluate the final response generated by the LLM generator in natural language form. These objectives encompass accuracy metrics, efficiency, and distribution of retrieval time (delay).

The methodology underpinning our approach is detailed in Algorithm \ref{alg3}. We have designed our system with a focus on three critical objectives to evaluate the performance of the retrieval methods comprehensively, enhancing the overall functionality of our RAG system. Each query's final response, generated by the LLM in natural language, is assessed according to these objectives, which include accuracy and efficiency metrics.

\textbf{Accuracy Metrics:}
For accuracy, we consider two key metrics: hit (\(h\), whether the response contains the correct answer) and recall (\(rc\), which assesses the system's ability to retrieve all relevant items). These metrics are crucial for assessing the precision and completeness of the system.

\textbf{Efficiency Metrics:}
Efficiency is evaluated based on the mean delay time (\(d_i\)) experienced by each retrieval method within the system. We utilize a distribution \(\sigma(d_i)\) to quantitatively represent each method's efficiency. This distribution helps the model understand the temporal performance across different retrieval strategies, ensuring the system delivers not only accurate but also timely responses.
\begin{equation}
    \sigma(d_i) = \frac{e^{1/d_{i}}}{\sum_{j=1}^K e^{1/d_{j}}},
\end{equation}
where methods with longer delays are assigned lower values, thus incentivizing quicker retrieval methods.

\textbf{Multi-Objective Optimization with GGI:}
To balance these objectives, we compute the multi-objective GGI value, which integrates accuracy and efficiency metrics, further detail of GGI property can be found in \cref{GGI-detail}. The GGI values for each objective are calculated as follows:
\begin{align}
    \label{loss-components}
    l_1 &= MSE(\max(f_\theta(x)), h), \quad &\text{(Loss- Accuracy - Hit)} \\
    l_2 &= MSE(\max(f_\theta(x)), rc), \quad &\text{(Loss-Accuracy - Recall)} \\
    l_3 &= KLDiv(f_\theta(x), \sigma(d_i)), \quad &\text{(Loss- Efficiency)}
\end{align}
Each loss component \( l_i \) corresponds to a specific objective:
\begin{itemize}[leftmargin=*]
    \item \( l_1 \) and \( l_2 \) measures the deviation in accuracy, encouraging the model to select a method with a high probability produce high recall and hit retrieval methods.
    \item \( l_3 \) quantifies the efficiency using the Kullback-Leibler Divergence (KLDiv) between the predicted arm selection distribution from the model and the efficiency distribution \( \sigma(d_i) \) encourage the model to select an efficient retrieval method.
\end{itemize}

The aggregate loss function to be minimized, representing the overall GGI, is then given by:
% \begin{equation}
% GGI_w (\mathbf{l_a}) = \sum_{i=1}^{D} w_i (l_a^i)_\sigma = \mathbf{w}^T (\mathbf{l}_a)_{\sigma},
% \end{equation}
% where \( w_1 > w_2 > \cdots > w_d > 0 \) and \( \sigma \) permutes the elements of \( \mathbf{l}_a \) such that \( (l_{\sigma})_i > (l_{\sigma})_{i+1} \). 
\begin{equation}
    \label{GGI-loss}
    Loss = GGI_w (\mathbf{l}) = \sum_{i=1}^{D} w_i (l_i)_\tau = \mathbf{w}^T (\textbf{l})_{\tau} 
\end{equation}
where \( w_1 > w_2 > \cdots > w_d > 0 \) and \( \tau \) permutes the elements of \( \mathbf{l} \) such that \( (l_i)_{\tau} > (l_{i+1})_{\tau} \).

In Equation \ref{GGI-loss}, the GGI function aggregates the individual loss components, weighted by \( w_i \), to update the parameter \(\theta\), optimizing towards a better response quality with satisfying user experience.

\section{Experiment}

    \subsection{Dataset \& Setup}

    \textbf{Datasets:} We evaluate our systems on two KGQA datasets WebQSP \cite{yih2016value} and ComplexWebQuestions (CWQ) \cite{talmor2018web} which contain up to 4-hop questions. The statistics of the datasets are given in \cref{tab:statistic}.

    \textbf{Baselines}: To valid the effectiveness of our MAB-enhanced KG-based RAG system under stationary environment, we compared it with state-of-the-art KG-based RAG systems, including the query language-based RAG: StructGPT \cite{structgpt}, ChatKBQA\cite{chatkbqa}, LLM agent-based RAG: Think-on-Graph \cite{thinkongraph} and Reason-on-Graph \cite{Rog}, and dense retrieval based RAG \cite{bge,decaf}.

    \textbf{Evaluation Metrics:} Following previous works, we evaluate the performance of our Retrieval-Augmented Generation system, all results are assessed based on the final generated response's hit rate and recall. We ran at least ten independent rounds with different seeds and reported the results as mean ± standard deviation to ensure the stability of our findings.

    \textbf{Implementations:}   We consistently use Llama-2-7b-chat-hf \cite{touvron2023llama} as the LLM generator, applying a standard RAG prompt \cite{llamaindex_prompt_engineering} across all methods to ensure a fair comparison.  All experiments are conducted on the Nvidia Tesla V100 graphical card with the Intel Xeon Platinum 8255C CPU. See \cref{app:imp} for detail set up.

    % 为了和其他的RAG 系统比较，我们会把他们描述成一个RAG系统，而不是一个retriever

\begin{figure}[ht!]
     \includegraphics[width=\linewidth]{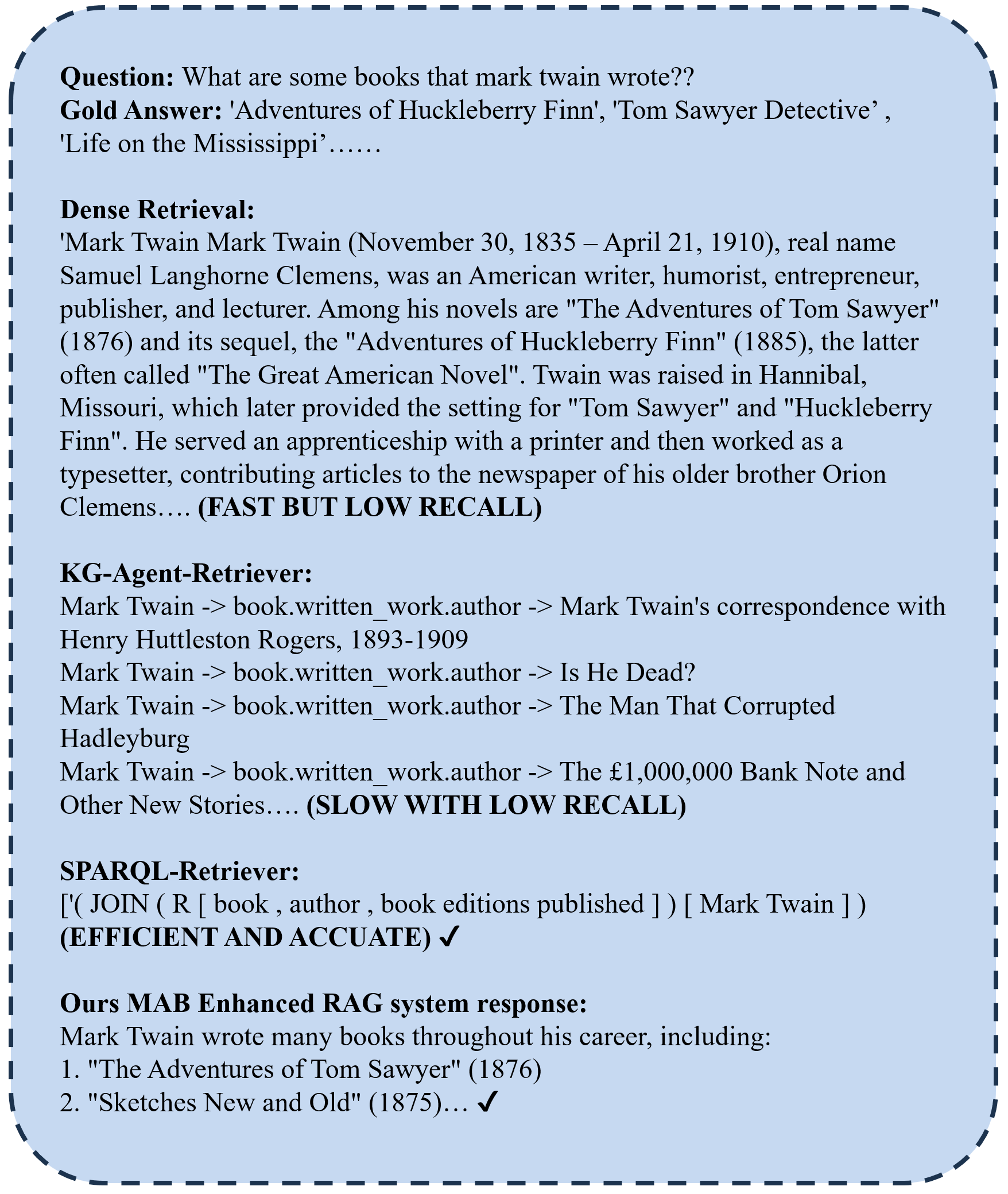}
     \caption{Comparison of retrieval methods for the query, "What are some books that Mark Twain wrote?" Dense Retrieval is fast but has low recall, while KG-Agent-Retriever provides broad coverage but is slow. Our system selects the SPARQL-Retriever \cite{chatkbqa}, which generates an accurate search language command for precise and efficient results.}
     \vspace{-5mm}
     \label{fig:Recall_case}
\end{figure}

\begin{table*}[ht!]
 \caption{Results under Stationary environment}
\centering
\small
\begin{tabular}{@{}cccccc@{}}
\toprule
\multirow{2}{*}{\textbf{Retriever Type}}           & \multirow{2}{*}{\textbf{Method}} & \multicolumn{2}{c}{\textbf{WebQSP}}      & \multicolumn{2}{c}{\textbf{CWQ}}             \\ 
                                            &                         & \textbf{Hit} $\uparrow$           & \textbf{Recall} $\uparrow$         & \textbf{Hit} $\uparrow$             & \textbf{Recall} $\uparrow$          \\ \midrule
\multirow{2}{*}{Dense   Retrieval}          & BGE \cite{bge}                     & 63.03          & 44.43          & 52.46            & 46.68            \\
                                            & DECAF \cite{decaf}                  & 71.37          & 50.93          & 47.46            & 41.47            \\ \midrule
\multirow{2}{*}{KG Query Languge Retrieval} & StructGPT \cite{structgpt}             & 75.56          & 55.26          & \textbackslash{} & \textbackslash{} \\
                                            & ChatKBQA \cite{chatkbqa}               & 80.77          & 64.31          & 77.37            & 69.46            \\ \midrule
\multirow{2}{*}{LLM agent Retreival}        & Think-on-graph \cite{thinkongraph}         & 66.64          & 47.24          & 58.90            & 52.49            \\
                                            & Reason-on-graph \cite{Rog}        & 85.70          & 75.07          & 56.63            & 52.38            \\ \midrule
\multicolumn{2}{c}{Ensemble  (DECAF+ChatKBQA+Reason-on-graph)}         & 83.74          & 67.52          & 67.93            & 68.01            \\ \midrule
\multirow{2}{*}{Static Router}              & LLM Router \cite{llamaIndexDefineSelector}             & 82.48          & 68.9           & 65.75            & 59.33            \\
                                            & NN-Router \cite{NN-router}              & 86.20          & 75.03          & 78.53            & 71.52            \\ \midrule
\textbf{Ours}              & GGI-MAB          & \textbf{86.64} & \textbf{75.60} & \textbf{79.35}   & \textbf{72.02}   \\ \bottomrule
\end{tabular}
% \vspace{-4mm}
\label{tab:station_comp}
\end{table*}

    \begin{table*}[ht!]
    \caption{Results under Non-stationary environment (mean ± std)}
\centering
\small
\begin{tabular}{@{}ccccc@{}}
\toprule
\textbf{Non-stationarity}                                                                    & \textbf{Method}      & \textbf{Test Hit } $\uparrow$    & \textbf{Test Recall} $\uparrow$  & \textbf{\begin{tabular}[c]{@{}c@{}}Test Retrieval Delay $\downarrow$\\       (second per query)\end{tabular}} \\ \midrule
\multirow{5}{*}{\begin{tabular}[c]{@{}c@{}}Retriever   \\      update\end{tabular}} & Retrieval Ensemble   & 83.74 ± 0.58          & 67.52 ± 0.77          & 15.00 ± 0.00                                                                                     \\
                                                                                             & Offline MO-MAB       & 82.25 ± 2.18          & 66.05 ± 2.56          & 13.32 ± 1.90                                                                                     \\
                                                                                             & NN Router \cite{NN-router}           & 81.48 ± 0.28          & 64.90 ± 0.23          & 14.09 ± 0.42                                                                                     \\
                                                                                             & LLM Router \cite{llamaIndexDefineSelector}          & 82.19 ± 0.47          & 67.11 ± 0.52          & 10.36 ± 3.98                                                                                     \\
                                                                                             & \textbf{Ours} & \textbf{84.80 ± 0.39} & \textbf{72.24 ± 0.71} & \textbf{5.88 ± 0.99}                                                                             \\ \midrule
\multirow{5}{*}{\begin{tabular}[c]{@{}c@{}}Domain\\      shift\end{tabular}}        & Retrieval Ensemble   & 67.93 ± 0.61          & 68.01± 0.34           & 15.00 ± 0.00                                                                                     \\
                                                                                             & Offline MO-MAB       & 64.76 ± 4.40          & 61.32 ± 2.90          & 7.78 ± 0.44                                                                            \\
                                                                                             & NN Router \cite{NN-router}            & 69.57 ± 4.52          & 63.99 ± 3.94          & \textbf{7.65 ± 1.81}                                                                             \\
                                                                                             & LLM Router \cite{llamaIndexDefineSelector}           & 65.75 ± 0.16          & 59.33 ± 0.25          & 9.39 ± 0.06                                                                                      \\
                                                                                             & \textbf{Ours} & \textbf{76.35 ± 0.69} & \textbf{69.47 ± 0.76} & 11.63 ± 0.28                                                                                     \\ \bottomrule
\end{tabular}
\vspace{-3mm}
\label{tab:non-station}
    \end{table*}

    \subsection{Research Questions and Main Results}
    % TODO: 我们的方法还很轻量，只需要简单的部署就可以实时运行。

    \textbf{ RQ1: Can our Multi-Armed Bandit enhanced Retrieval-Augmented Generation system effectively improve performance compared to RAG systems that rely on a single retrieval method?}
    
    The comparative analysis, summarized in \cref{tab:station_comp}, revealed that our MAB-enhanced RAG system demonstrated superior performance across both datasets. Notably, on the CWQ dataset, which poses more intricate multi-hop reasoning challenges, our method exceeds the next-best performance by nearly 2\% in hit rate and over 2.5\% in recall.

\begin{figure}[ht!]
     \includegraphics[width=\linewidth]{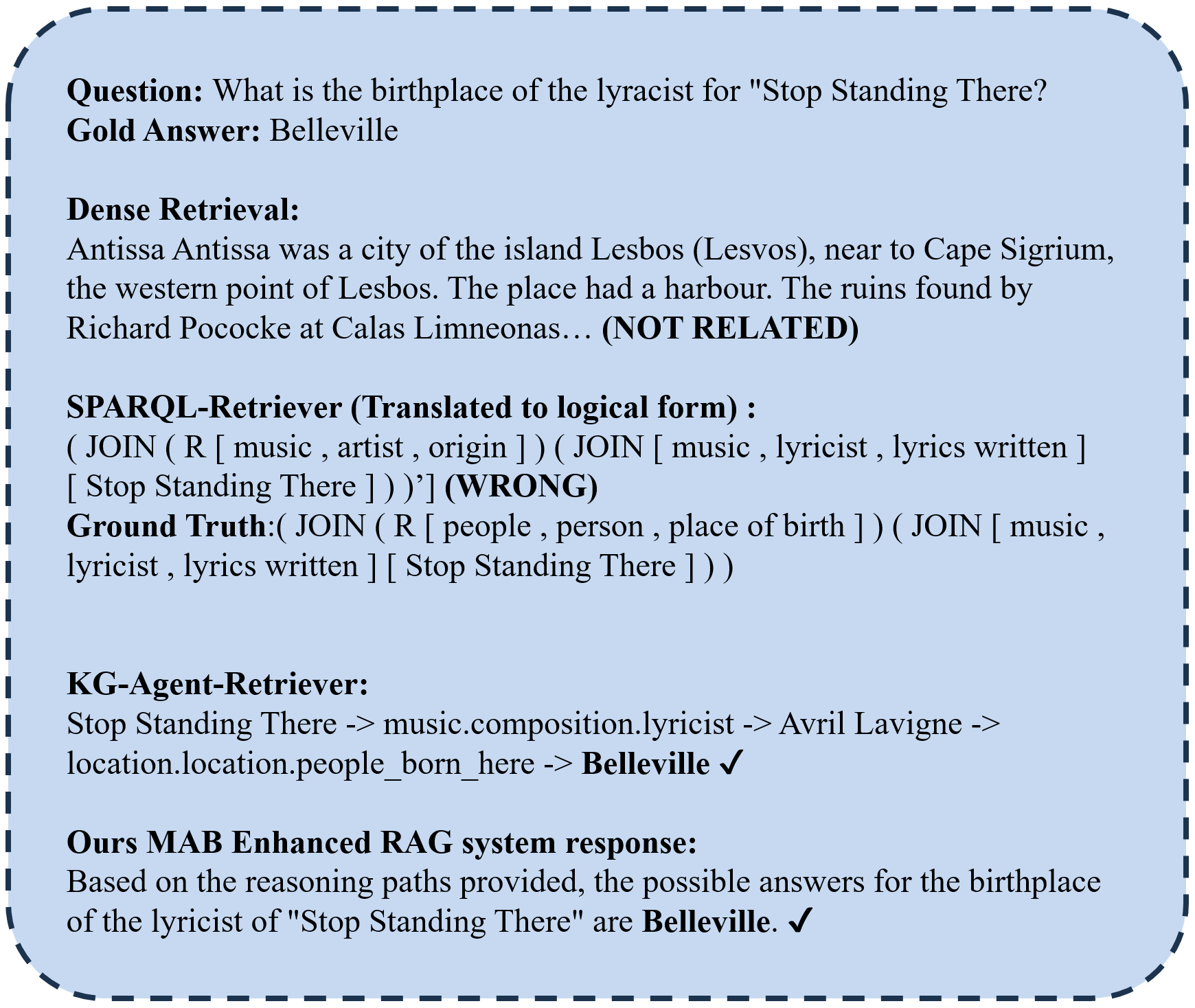}
     \vspace{-2mm}
     \caption{KG Agent gives the most accurate context}
     \label{fig:example_cwq}
\end{figure}

    We present examples of our MAB-enhanced RAG systems superior case. 
    In the first case \cref{fig:example_cwq} derived from the challenging CWQ dataset, the query pertains to the birthplace of the lyricist for "Stop Standing There." Dense retrieval fails to relate the query to relevant information, and while the SPARQL retriever approaches a correct formulation, it ultimately generates a wrong query language. Thanks to the reasoning ability of LLM, the LLM-based KG agent successfully retrieves the related triplets from the knowledge graph and enables our MAB Enhanced RAG system to give an accurate response. 
    
    In the second case \cref{fig:Recall_case}, from the WebQSP dataset where the user queries, "What are some books that Mark Twain wrote?" This question is challenging in terms of achieving high recall since all retrieval methods can provide related context, but not all can accurately list the books.
    Our MAB-enhanced RAG system effectively selects the appropriate methods (SPARQL generator) to achieve the highest recall, significantly outperforming individual retrieval approaches.

    Our MAB-enhanced RAG system, effectively optimizes the selection process of retrieval methods, thereby proving to be highly effective in improving overall system performance. Furthermore, as illustrated in \nameref{variants}, we evaluate our method across different Large Language Model generators to prove the robustness of our system.

    \begin{table*}[t!]
    \caption{Results of proposed multi-objective MAB algorithm under station environments (mean ± std)}
    \centering
    \small
\begin{tabular}{@{}ccccc@{}}
\toprule
\multicolumn{2}{c}{\textbf{Method}}            & \textbf{Test Hit} $\uparrow$     & \textbf{Test Recall} $\uparrow$  & \textbf{\begin{tabular}[c]{@{}c@{}}Test Retrieval Delay $\downarrow$\\       (second per query)\end{tabular}} \\ \midrule
\multirow{5}{*}{Baselines} & UCB \cite{UCB}               & 78.44   ± 6.07        & 62.18 ± 10.08         & 5.80 ± 6.08                                                                                        \\
                           & Thompsom Sampling \cite{Thompsonsampling} & 84.12   ± 2.20        & 71.82 ± 4.93          & 6.60 ± 5.50                                                                                        \\
                           & LinUCB \cite{linucb}            & 81.99 ±   2.91        & 68.55 ± 5.30          & 5.31 ± 2.80                                                                                        \\
                           & SO-Deep-MAB \cite{DeepMAB}       & 86.79 ± 0.33          & 75.18 ± 0.18          & 11.1 ± 0.39                                                                                        \\
                           & MOU-UCB\cite{wanigasekara2019learning}           & 85.55 ± 0.89          & 75.05 ± 0.15          & 5.52 ± 1.35                                                                                        \\ \midrule
\multirow{2}{*}{Ours}      & MO-MAB       & 85.31 ± 0.55          & 74.38 ± 0.48         & 5.20 ± 0.58                                                                                        \\
                           & GGI-MO-MAB        & \textbf{86.64 ± 0.29} & \textbf{75.60 ± 0.38} & \textbf{4.84 ± 0.81}                                                                               \\ \bottomrule
\end{tabular}
\vspace{-3mm}
\label{tab:mo-results}
    \end{table*}
    
    \textbf{RQ2: Can the MAB enhanced RAG system adapts dynamically to the non-stationary nature of real-world environments, ensuring that they continuously meet evolving query demands and operational conditions?}

    To evaluate our methods under non-stationary environments, we use two non-stationary settings: (1) we employed the KG agent-based retrieval method \cite{thinkongraph} during the training phase. For online testing, we switched to  \cite{Rog} a method with superior performance, to simulate the effect of upgrading backend retrievers independently to enhance system functionality. This approach tests the system's ability to adapt seamlessly to improvements in retrieval methods, reflecting real-world conditions where continuous updates are crucial for maintaining system efficacy. (2) To simulate the shift in query domains resulting from changes in trending topics, we initially train our methods using the WebQSP dataset. Subsequently, we evaluate the system's adaptability by testing it on the ComplexWebQuestions dataset. This approach allows us to assess how well the system can handle transitions between different types of query complexities and content, mirroring real-world scenarios where query characteristics can vary significantly due to external influences.

    The results, as detailed in \cref{tab:non-station}, in the first scenario, during the retriever upgrade tests, our method demonstrated the highest Test Hit and Test Recall rates f 84.80\% and 72.24\% respectively, with a significantly reduced Test Retrieval Delay of 5.88 seconds per query. This improvement stems from our system's capability to leverage partial information during testing to continuously refine the model. In contrast, retrieval ensemble methods, which require running all retrieval methods, struggle with denoising information from different structures of retrieval results leading to the longest Retrieval delay. Both the offline classifier and offline multi-objective MAB (MO-MAB) were unable to adapt to the upgrades, resulting in inferior performance.

    In the second scenario, our approach effectively adapted to domain shifts by utilizing the slower KG agent retriever and SPARQL generator retrieval methods. Although it needs more retrieval time at 11.63 seconds per query compared to some offline methods, it significantly outperformed comparative methods in accuracy metrics, achieving a Test Hit rate of 76.35\% and a Test Recall of 69.47\%. 
    
    The results further highlight our system's capacity to dynamically adjust operational parameters in response to evolving query complexities, ensuring high-quality user interactions even under challenging conditions.

    % For the second scenario, our approach learn to adapt this domain shift and start to use KG agent retriever and SPARQL generator which are much more slower but again outperformed the comparative methods in accuracy matrics, achieving a Test Hit rate of 76.35\% and a Test Recall of 69.47\%. While the Test Retrieval Delay was slightly higher than some of the offline methods, it remained competitive at 11.63 seconds per query. These results affirm the system's capacity to effectively manage transitions in query domains influenced by external social changes, thus ensuring consistent performance in real-world applications.
    % The results underscore our system’s capability to dynamically adjust its operational parameters in response to evolving query complexities, ensuring that the quality of user interactions remains high even under demanding conditions.

    \textbf{RQ3: How can the Generalized Gini Index be effectively utilized to balance multiple performance metrics in RAG systems}

    In \cref{tab:mo-results} we evaluate our proposed method on the WebQSP dataset, results highlight the effectiveness of the Generalized Gini Index enhanced Multi-Objective Multi-Armed Bandit (GGI-MO-MAB), achieving the highest Test Hit rate and Test Recall, while maintaining the lowest retrieval delay compared to the baselines.
    Non-contextual baselines like UCB \cite{UCB} and Thompson Sampling \cite{Thompsonsampling} approximate only a single optimal retrieval method.
    LinUCB \cite{linucb} under-performs due to its inability to handle the high-dimensional, complex natural language embeddings.
    Single-objective deep contextual MAB models, while improving accuracy metrics such as Hit rate, often neglect retrieval time, adversely affecting user experience.
    Our GGI-MO-MAB can also outperform multi-objective baseline MOU-UCB \cite{wanigasekara2019learning}.
    To underscore the efficacy of our approach, we include an ablation study comparing the GGI function to a learnable weight aggregation baseline (MO-MAB), confirming the robust performance improvement of our method.

    \textbf{RQ4: What are the effects of implementing multiple retrieval methods, such as dense retrieval and KG agent retrieval methods, on the response times and accuracy under different real-world scenarios?}
    \label{RQ1}
    
    \begin{figure}[ht!]
     \includegraphics[width=\linewidth]{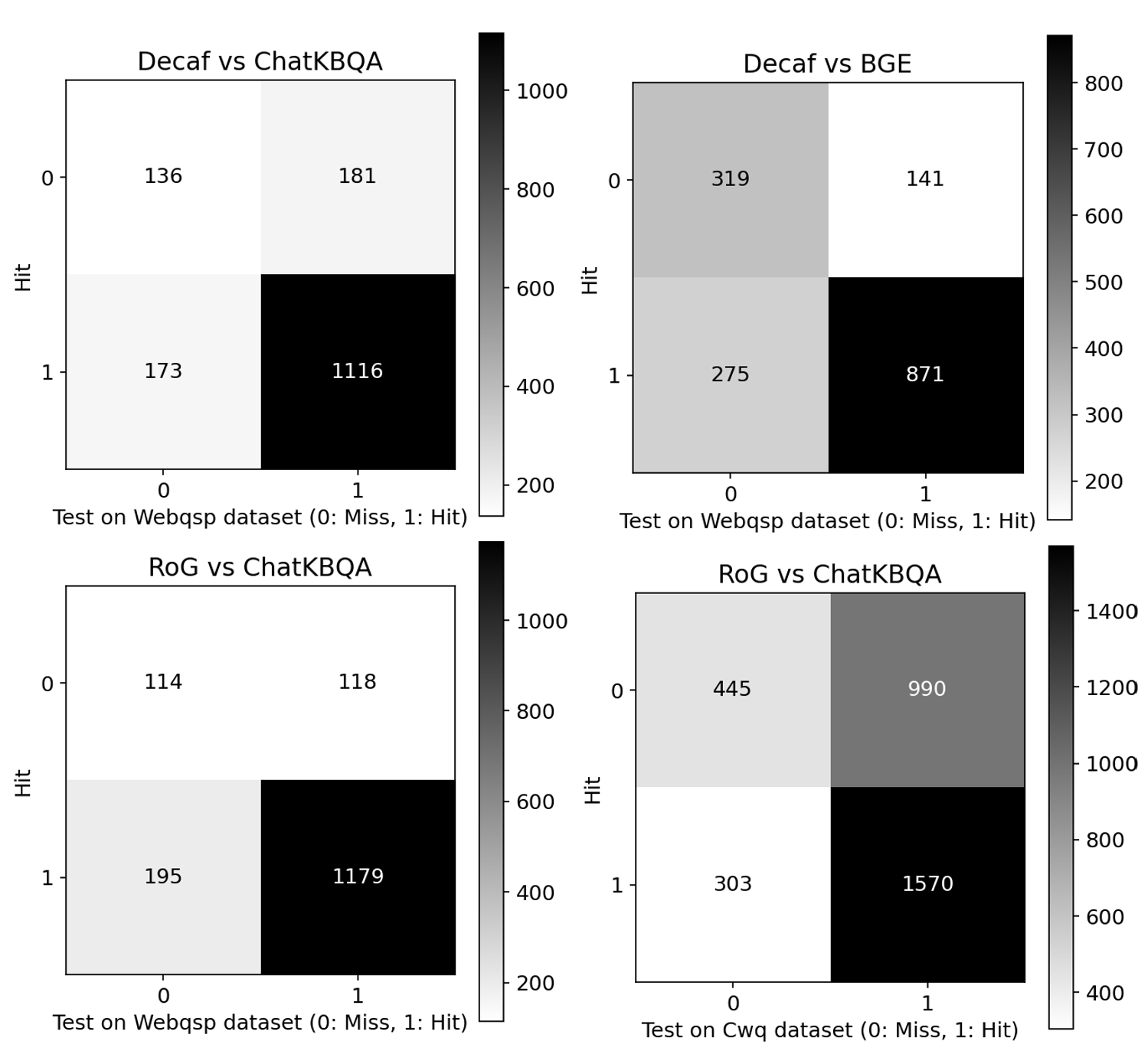}
     \caption{Confusion matrices comparing retrieval methods (Decaf, ChatKBQA, BGE, RoG) on WebQSP and CWQ datasets, indicating distinctiveness among methods.}
     \vspace{-3mm}
     \label{confusion_matrix}
    \end{figure}
    The comparison of different types of retrieval methods is shown in \cref{confusion_matrix}, we also employed the Jaccard Similarity Coefficient to assess the Hit metric across results. Our findings reveal an average coefficient of 0.738, with the lowest observed at 0.496, indicating the distinctiveness of the results obtained by different retrieval strategies. 
    
    Moreover, while methods such as ChatKBQA and Reason-on-graph showed strong results on WebQSP, they were less effective on the more challenging CWQ dataset, highlighting the importance of retrieval method selection based on the complexity and nature of the dataset. Our system's consistent performance across different datasets underscores its robustness and adaptability, making it particularly suitable for diverse real-world applications where query demands and operational conditions can vary significantly.
    
    In terms of response time, we observed significant differences in processing time; for instance, dense-vector retrieval methods average around 1 second, whereas more complex methods like ChatKBQA \cite{chatkbqa}, due to multiple interactions with ChatGPT \cite{ChatGPT} to get an executable query code, can take 15-30 seconds. These findings highlight the trade-off between complexity and efficiency in retrieval operations.

    Our findings confirm that the choice of retrieval method significantly impacts the accuracy and efficiency of RAG systems.

    %With respect to the effects of routine within multiple retrieval methods

\section{Related Work}

\textbf{KG-based RAG Systems:} Retrieval-Augmented Generation (RAG)\cite{RAG} mitigates the hallucination issue of LLMs by retrieving external knowledge to enhance the accuracy and reliability of generation content. Recent RAG advancements have increasingly incorporated Knowledge Graphs (KGs) \cite{Rog,thinkongraph,KAPING,G-Retriever}, which store structured factual information, enabling more systematic reasoning by LLMs \cite{pan2024unifying}. KGs support diverse retrieval methods, each with different capabilities and costs, as detailed in \cref{Retrieval-methods-on-KG}.

Our analysis of retrieval methods, discussed in \cref{RQ1}, shows that current KG-based RAG systems\cite{Rog,thinkongraph,KAPING,G-Retriever} predominantly rely on a single retrieval method, which often fails to meet the varied demands of real-world applications. These systems typically assume a stationary environment and remain static without subsequent fine-tuning, making them unable to adapt to potential shifts in the query domain and upgrades of the backend retriever. To address these issues,our work aims to develop an MAB-enhanced RAG system that strategically combines multiple retrievers. By leveraging real-time feedback, our system can dynamically adjust retrieval strategies to meet the evolving demands of diverse application scenarios of the RAG system effectively.

To our knowledge, the concurrent research by \cite{sawarkar2024blended} is one of the few studies attempting to integrate multiple retrieval methods, but it focuses on textual data sources and lacks the continuous optimization crucial for RAG systems in non-stationary environments. 

% The Multi-Armed Bandit (MAB) \cite{katehakis1987multi} framework is designed to optimize the balance between exploiting historical data and exploring new information. This framework is divided into two main categories: context-free \cite{bubeck2012regret}, without external information, and contextual bandits \cite{contexualMAB}, which incorporate context like user feature. While traditional contextual bandits typically assume a linear relationship between context and expected rewards \cite{pmlr-v19-slivkins11a}, recent developments have introduced non-linear models through deep learning \cite{DeepMAB,zhou2020neural,shi2023deep}.

% In particular, the multi-objective contextual multi-armed bandit (MOCMAB) algorithm \cite{dominantMOMAB}  maximize total rewards across several objectives, managing both dominant and non-dominant objectives. Furthermore, some approaches \cite{MOMABGGI,musicMOMAB} utilize the Generalized Gini Index (GGI) to convert multi-objective challenges into single-objective optimizations, simplifying the decision-making process in dynamic environments.

\textbf{Multi-Armed Bandit Algorithms:} The Multi-Armed Bandit (MAB) \cite{katehakis1987multi} framework optimizes the balance between exploiting historical data and exploring new information. It includes two main types: context-free \cite{bubeck2012regret}, which operates without external information, and contextual bandits \cite{contexualMAB}, which incorporate contextual data such as user features. Traditional contextual bandits assume a linear relationship between context and expected rewards \cite{pmlr-v19-slivkins11a}, but recent developments have introduced non-linear models through deep learning \cite{DeepMAB,zhou2020neural,shi2023deep}. The multi-objective contextual MAB (MOCMAB) algorithm \cite{dominantMOMAB} maximizes rewards across multiple objectives, managing both dominant and non-dominant goals. Some approaches \cite{MOMABGGI,musicMOMAB} use the Generalized Gini Index (GGI) to convert multi-objective challenges into single-objective optimizations, simplifying decision-making in dynamic environments.

However, existing contextual bandit algorithms often assume reward is linear with respect to the context feature \cite{dominantMOMAB,musicMOMAB,linucb}, limiting their representational capacity to match user query patterns with retrieval strategies effectively, or they focus solely on single-objective optimization \cite{DeepMAB,zhou2020neural,shi2023deep}, which does not suffice for complex RAG systems with requirements of performance and real-time limitation. Therefore, in this work, we adopt a non-linear multi-objective contextual MAB model.

\section{Conclusion}

In this work, we introduced a novel KG-based RAG framework enhanced by a Multi-Armed Bandit (MAB) model. By leveraging real-time user feedback, our system dynamically adapts to shifting query demands and backend upgrades. We further incorporated the Generalized Gini Index to balance multiple objectives, ensuring that the system delivers both informative and timely responses.

Our comprehensive evaluations on two well-established KBQA datasets, WebQuestionSP and ComplexWebQuestions, demonstrate that our approach not only significantly outperforms baseline methods in non-stationary environments but also surpasses state-of-the-art KG-based RAG systems in stationary settings. These results underscore the robustness, adaptability, and practical applicability of our framework in real-world scenarios where query demands and operational conditions are constantly evolving.

% Moreover, with a focus on user experience, we implemented a Generalized Gini Index-based multi-objective learning algorithm. This algorithm ensures superior response quality by balancing crucial performance metrics with retrieval time, thus addressing the practical challenges faced during deployment.

% Extensive testing on benchmark KGQA datasets has proven that our approach not only sustains robustness in the face of environmental disturbances but also achieves state-of-the-art performance in more stable conditions. This validates our model's effectiveness in enhancing the reasoning capabilities of large language models through strategic retrieval method selection and adaptive learning.

\bibliography{aaai25}

\begin{thebibliography}{60}
\providecommand{\natexlab}[1]{#1}

\bibitem[{lla(2024)}]{llamaIndexDefineSelector}
 2024.
\newblock Define Selector Module for Routing --- Llama Index Documentation.
\newblock \url{https://docs.llamaindex.ai/en/stable/examples/retrievers/router_retriever/#define-selector-module-for-routing}.
\newblock Accessed: 2024-08-07.

\bibitem[{Agrawal and Goyal(2013)}]{Thompsonsampling}
Agrawal, S.; and Goyal, N. 2013.
\newblock Thompson sampling for contextual bandits with linear payoffs.
\newblock In \emph{International conference on machine learning}, 127--135. PMLR.

\bibitem[{Alan, Ayd{\i}n, and Karaarslan(2024)}]{alan2024rag}
Alan, A.~Y.; Ayd{\i}n, {\"O}.; and Karaarslan, E. 2024.
\newblock A RAG-based Question Answering System Proposal for Understanding Islam: MufassirQAS LLM.
\newblock \emph{Available at SSRN 4707470}.

\bibitem[{Auer(2002)}]{UCB}
Auer, P. 2002.
\newblock Using confidence bounds for exploitation-exploration trade-offs.
\newblock \emph{Journal of Machine Learning Research}, 3(Nov): 397--422.

\bibitem[{Auer, Cesa-Bianchi, and Fischer(2002)}]{auer2002finite}
Auer, P.; Cesa-Bianchi, N.; and Fischer, P. 2002.
\newblock Finite-time analysis of the multiarmed bandit problem.
\newblock \emph{Machine learning}, 47: 235--256.

\bibitem[{Baek, Aji, and Saffari(2023)}]{baek2023knowledge}
Baek, J.; Aji, A.~F.; and Saffari, A. 2023.
\newblock Knowledge-augmented language model prompting for zero-shot knowledge graph question answering.
\newblock \emph{arXiv preprint arXiv:2306.04136}.

\bibitem[{Bang et~al.(2023)Bang, Cahyawijaya, Lee, Dai, Su, Wilie, Lovenia, Ji, Yu, Chung et~al.}]{bang2023multitask}
Bang, Y.; Cahyawijaya, S.; Lee, N.; Dai, W.; Su, D.; Wilie, B.; Lovenia, H.; Ji, Z.; Yu, T.; Chung, W.; et~al. 2023.
\newblock A multitask, multilingual, multimodal evaluation of chatgpt on reasoning, hallucination, and interactivity.
\newblock \emph{arXiv preprint arXiv:2302.04023}.

\bibitem[{BehnamGhader, Miret, and Reddy(2023)}]{behnamghader2023can}
BehnamGhader, P.; Miret, S.; and Reddy, S. 2023.
\newblock Can Retriever-Augmented Language Models Reason? The Blame Game Between the Retriever and the Language Model.
\newblock In \emph{Findings of the Association for Computational Linguistics: EMNLP 2023}, 15492--15509.

\bibitem[{Brown et~al.(2020)Brown, Mann, Ryder, Subbiah, Kaplan, Dhariwal, Neelakantan, Shyam, Sastry, Askell et~al.}]{brown2020language}
Brown, T.; Mann, B.; Ryder, N.; Subbiah, M.; Kaplan, J.~D.; Dhariwal, P.; Neelakantan, A.; Shyam, P.; Sastry, G.; Askell, A.; et~al. 2020.
\newblock Language models are few-shot learners.
\newblock \emph{Advances in neural information processing systems}, 33: 1877--1901.

\bibitem[{Bubeck, Cesa-Bianchi et~al.(2012)}]{bubeck2012regret}
Bubeck, S.; Cesa-Bianchi, N.; et~al. 2012.
\newblock Regret analysis of stochastic and nonstochastic multi-armed bandit problems.
\newblock \emph{Foundations and Trends{\textregistered} in Machine Learning}, 5(1): 1--122.

\bibitem[{Busa-Fekete et~al.(2017)Busa-Fekete, Sz{\"o}r{\'e}nyi, Weng, and Mannor}]{MOMABGGI}
Busa-Fekete, R.; Sz{\"o}r{\'e}nyi, B.; Weng, P.; and Mannor, S. 2017.
\newblock Multi-objective bandits: Optimizing the generalized Gini index.
\newblock In \emph{International Conference on Machine Learning}, 625--634. PMLR.

\bibitem[{Chen et~al.()Chen, Guo, Chen, Ma, Zeng, Liao, Zhang, and Xie}]{chentraining}
Chen, C.; Guo, C.; Chen, R.; Ma, G.; Zeng, M.; Liao, X.; Zhang, X.; and Xie, S. ????
\newblock Training for Stable Explanation for Free.
\newblock In \emph{The Thirty-eighth Annual Conference on Neural Information Processing Systems}.

\bibitem[{Chowdhery et~al.(2023)Chowdhery, Narang, Devlin, Bosma, Mishra, Roberts, Barham, Chung, Sutton, Gehrmann et~al.}]{chowdhery2023palm}
Chowdhery, A.; Narang, S.; Devlin, J.; Bosma, M.; Mishra, G.; Roberts, A.; Barham, P.; Chung, H.~W.; Sutton, C.; Gehrmann, S.; et~al. 2023.
\newblock Palm: Scaling language modeling with pathways.
\newblock \emph{Journal of Machine Learning Research}, 24(240): 1--113.

\bibitem[{Collier and Llorens(2018)}]{DeepMAB}
Collier, M.; and Llorens, H.~U. 2018.
\newblock Deep contextual multi-armed bandits.
\newblock \emph{arXiv preprint arXiv:1807.09809}.

\bibitem[{Devlin et~al.(2018)Devlin, Chang, Lee, and Toutanova}]{bert}
Devlin, J.; Chang, M.-W.; Lee, K.; and Toutanova, K. 2018.
\newblock Bert: Pre-training of deep bidirectional transformers for language understanding.
\newblock \emph{arXiv preprint arXiv:1810.04805}.

\bibitem[{Du et~al.(2024)Du, Chen, Zhang, Ma, and Liu}]{Molecule}
Du, W.; Chen, J.; Zhang, X.; Ma, Z.; and Liu, S. 2024.
\newblock Molecule joint auto-encoding: trajectory pretraining with 2D and 3D diffusion.
\newblock In \emph{Proceedings of the 37th International Conference on Neural Information Processing Systems}, NIPS '23. Curran Associates Inc.

\bibitem[{Du et~al.(2021)Du, Qian, Liu, Ding, Qiu, Yang, and Tang}]{du2021glm}
Du, Z.; Qian, Y.; Liu, X.; Ding, M.; Qiu, J.; Yang, Z.; and Tang, J. 2021.
\newblock Glm: General language model pretraining with autoregressive blank infilling.
\newblock \emph{arXiv preprint arXiv:2103.10360}.

\bibitem[{Ennen et~al.(2023)Ennen, Freddi, Lin, Kung, Wang, Yang, Shiu, and Bernacchia}]{ennen2023hierarchical}
Ennen, P.; Freddi, F.; Lin, C.-J.; Kung, P.-N.; Wang, R.; Yang, C.-Y.; Shiu, D.-s.; and Bernacchia, A. 2023.
\newblock Hierarchical Representations in Dense Passage Retrieval for Question-Answering.
\newblock In \emph{Proceedings of the Sixth Fact Extraction and VERification Workshop (FEVER)}, 17--28.

\bibitem[{Gamage et~al.(2024)Gamage, Mills, De~Silva, Manic, Moraliyage, Jennings, and Alahakoon}]{gamage2024multi}
Gamage, G.; Mills, N.; De~Silva, D.; Manic, M.; Moraliyage, H.; Jennings, A.; and Alahakoon, D. 2024.
\newblock Multi-Agent RAG Chatbot Architecture for Decision Support in Net-Zero Emission Energy Systems.
\newblock In \emph{2024 IEEE International Conference on Industrial Technology (ICIT)}, 1--6. IEEE.

\bibitem[{Guo et~al.(2024)Guo, Zou, Chen, Qu, Chi, Yu, and Chang}]{guo2024sample}
Guo, S.; Zou, L.; Chen, H.; Qu, B.; Chi, H.; Yu, P.~S.; and Chang, Y. 2024.
\newblock Sample Efficient Offline-to-Online Reinforcement Learning.
\newblock \emph{IEEE Transactions on Knowledge and Data Engineering}, 36(3): 1299--1310.

\bibitem[{He et~al.(2024)He, Tian, Sun, Chawla, Laurent, LeCun, Bresson, and Hooi}]{G-Retriever}
He, X.; Tian, Y.; Sun, Y.; Chawla, N.~V.; Laurent, T.; LeCun, Y.; Bresson, X.; and Hooi, B. 2024.
\newblock G-Retriever: Retrieval-Augmented Generation for Textual Graph Understanding and Question Answering.
\newblock \emph{arXiv preprint arXiv:2402.07630}.

\bibitem[{Jenkins(2017)}]{jenkins2017measurement}
Jenkins, S. 2017.
\newblock The measurement of income inequality.
\newblock In \emph{Economic inequality and poverty}, 3--38. Routledge.

\bibitem[{Ji et~al.(2023)Ji, Lee, Frieske, Yu, Su, Xu, Ishii, Bang, Madotto, and Fung}]{hallucination}
Ji, Z.; Lee, N.; Frieske, R.; Yu, T.; Su, D.; Xu, Y.; Ishii, E.; Bang, Y.~J.; Madotto, A.; and Fung, P. 2023.
\newblock Survey of hallucination in natural language generation.
\newblock \emph{ACM Computing Surveys}, 55(12): 1--38.

\bibitem[{Jiang et~al.(2023{\natexlab{a}})Jiang, Sablayrolles, Mensch, Bamford, Chaplot, Casas, Bressand, Lengyel, Lample, Saulnier et~al.}]{jiang2023mistral}
Jiang, A.~Q.; Sablayrolles, A.; Mensch, A.; Bamford, C.; Chaplot, D.~S.; Casas, D. d.~l.; Bressand, F.; Lengyel, G.; Lample, G.; Saulnier, L.; et~al. 2023{\natexlab{a}}.
\newblock Mistral 7B.
\newblock \emph{arXiv preprint arXiv:2310.06825}.

\bibitem[{Jiang et~al.(2023{\natexlab{b}})Jiang, Zhou, Dong, Ye, Zhao, and Wen}]{structgpt}
Jiang, J.; Zhou, K.; Dong, Z.; Ye, K.; Zhao, W.~X.; and Wen, J.-R. 2023{\natexlab{b}}.
\newblock Structgpt: A general framework for large language model to reason over structured data.
\newblock \emph{arXiv preprint arXiv:2305.09645}.

\bibitem[{Karpukhin et~al.(2020)Karpukhin, Oguz, Min, Lewis, Wu, Edunov, Chen, and Yih}]{DPR}
Karpukhin, V.; Oguz, B.; Min, S.; Lewis, P.~S.; Wu, L.; Edunov, S.; Chen, D.; and Yih, W.-t. 2020.
\newblock Dense Passage Retrieval for Open-Domain Question Answering.
\newblock In \emph{EMNLP (1)}, 6769--6781.

\bibitem[{Katehakis and Veinott~Jr(1987)}]{katehakis1987multi}
Katehakis, M.~N.; and Veinott~Jr, A.~F. 1987.
\newblock The multi-armed bandit problem: decomposition and computation.
\newblock \emph{Mathematics of Operations Research}, 12(2): 262--268.

\bibitem[{Langford and Zhang(2007)}]{epsilon-greedy}
Langford, J.; and Zhang, T. 2007.
\newblock The epoch-greedy algorithm for multi-armed bandits with side information.
\newblock \emph{Advances in neural information processing systems}, 20.

\bibitem[{Lee et~al.(2022)Lee, Seo, Lee, Abbeel, and Shin}]{lee2022offline}
Lee, S.; Seo, Y.; Lee, K.; Abbeel, P.; and Shin, J. 2022.
\newblock Offline-to-online reinforcement learning via balanced replay and pessimistic q-ensemble.
\newblock In \emph{Conference on Robot Learning}, 1702--1712. PMLR.

\bibitem[{Lewis et~al.(2020)Lewis, Perez, Piktus, Petroni, Karpukhin, Goyal, K{\"u}ttler, Lewis, Yih, Rockt{\"a}schel et~al.}]{RAG}
Lewis, P.; Perez, E.; Piktus, A.; Petroni, F.; Karpukhin, V.; Goyal, N.; K{\"u}ttler, H.; Lewis, M.; Yih, W.-t.; Rockt{\"a}schel, T.; et~al. 2020.
\newblock Retrieval-augmented generation for knowledge-intensive nlp tasks.
\newblock \emph{Advances in Neural Information Processing Systems}, 33: 9459--9474.

\bibitem[{Li et~al.(2010)Li, Chu, Langford, and Schapire}]{linucb}
Li, L.; Chu, W.; Langford, J.; and Schapire, R.~E. 2010.
\newblock A contextual-bandit approach to personalized news article recommendation.
\newblock In \emph{Proceedings of the 19th international conference on World wide web}, 661--670.

\bibitem[{LlamaIndex(2024)}]{llamaindex_prompt_engineering}
LlamaIndex. 2024.
\newblock Prompt Engineering for RAG.
\newblock Accessed: 2024-05-16.

\bibitem[{Luo et~al.(2023{\natexlab{a}})Luo, Tang, Peng, Guo, Zhang, Ma, Dong, Song, Lin et~al.}]{chatkbqa}
Luo, H.; Tang, Z.; Peng, S.; Guo, Y.; Zhang, W.; Ma, C.; Dong, G.; Song, M.; Lin, W.; et~al. 2023{\natexlab{a}}.
\newblock Chatkbqa: A generate-then-retrieve framework for knowledge base question answering with fine-tuned large language models.
\newblock \emph{arXiv preprint arXiv:2310.08975}.

\bibitem[{Luo et~al.(2023{\natexlab{b}})Luo, Ju, Xiong, Li, Haffari, and Pan}]{luo2023chatrule}
Luo, L.; Ju, J.; Xiong, B.; Li, Y.-F.; Haffari, G.; and Pan, S. 2023{\natexlab{b}}.
\newblock Chatrule: Mining logical rules with large language models for knowledge graph reasoning.
\newblock \emph{arXiv preprint arXiv:2309.01538}.

\bibitem[{Luo et~al.(2023{\natexlab{c}})Luo, Li, Haffari, and Pan}]{Rog}
Luo, L.; Li, Y.-F.; Haffari, G.; and Pan, S. 2023{\natexlab{c}}.
\newblock Reasoning on graphs: Faithful and interpretable large language model reasoning.
\newblock \emph{arXiv preprint arXiv:2310.01061}.

\bibitem[{Mahajan and Teneketzis(2008)}]{contexualMAB}
Mahajan, A.; and Teneketzis, D. 2008.
\newblock Multi-armed bandit problems.
\newblock In \emph{Foundations and applications of sensor management}, 121--151. Springer.

\bibitem[{Mehrotra, Xue, and Lalmas(2020)}]{musicMOMAB}
Mehrotra, R.; Xue, N.; and Lalmas, M. 2020.
\newblock Bandit based optimization of multiple objectives on a music streaming platform.
\newblock In \emph{Proceedings of the 26th ACM SIGKDD international conference on knowledge discovery \& data mining}, 3224--3233.

\bibitem[{OpenAI(2024)}]{ChatGPT}
OpenAI. 2024.
\newblock ChatGPT.
\newblock \url{https://openai.com/chatgpt}.
\newblock Accessed: 2024-05-20.

\bibitem[{OpenAI(2023)}]{openai2023gpt}
OpenAI, R. 2023.
\newblock Gpt-4 technical report. arxiv 2303.08774.
\newblock \emph{View in Article}, 2(5).

\bibitem[{Pan et~al.(2024)Pan, Luo, Wang, Chen, Wang, and Wu}]{pan2024unifying}
Pan, S.; Luo, L.; Wang, Y.; Chen, C.; Wang, J.; and Wu, X. 2024.
\newblock Unifying large language models and knowledge graphs: A roadmap.
\newblock \emph{IEEE Transactions on Knowledge and Data Engineering}.

\bibitem[{Petroni et~al.(2020)Petroni, Piktus, Fan, Lewis, Yazdani, De~Cao, Thorne, Jernite, Karpukhin, Maillard et~al.}]{petroni2020kilt}
Petroni, F.; Piktus, A.; Fan, A.; Lewis, P.; Yazdani, M.; De~Cao, N.; Thorne, J.; Jernite, Y.; Karpukhin, V.; Maillard, J.; et~al. 2020.
\newblock KILT: a benchmark for knowledge intensive language tasks.
\newblock \emph{arXiv preprint arXiv:2009.02252}.

\bibitem[{Reis et~al.(2019)Reis, Rocha, Phan, Griffin, Le, and Rio}]{NN-router}
Reis, J.; Rocha, M.; Phan, T.~K.; Griffin, D.; Le, F.; and Rio, M. 2019.
\newblock Deep Neural Networks for Network Routing.
\newblock In \emph{2019 International Joint Conference on Neural Networks (IJCNN)}, 1--8.

\bibitem[{Sanh et~al.(2019)Sanh, Debut, Chaumond, and Wolf}]{distilbert}
Sanh, V.; Debut, L.; Chaumond, J.; and Wolf, T. 2019.
\newblock DistilBERT, a distilled version of BERT: smaller, faster, cheaper and lighter.
\newblock \emph{arXiv preprint arXiv:1910.01108}.

\bibitem[{Sawarkar, Mangal, and Solanki(2024)}]{sawarkar2024blended}
Sawarkar, K.; Mangal, A.; and Solanki, S.~R. 2024.
\newblock Blended RAG: Improving RAG (Retriever-Augmented Generation) Accuracy with Semantic Search and Hybrid Query-Based Retrievers.
\newblock \emph{arXiv preprint arXiv:2404.07220}.

\bibitem[{Shi et~al.(2023)Shi, Xiao, Pickard, Chen, and Chen}]{shi2023deep}
Shi, Q.; Xiao, F.; Pickard, D.; Chen, I.; and Chen, L. 2023.
\newblock Deep neural network with linucb: A contextual bandit approach for personalized recommendation.
\newblock In \emph{Companion Proceedings of the ACM Web Conference 2023}, 778--782.

\bibitem[{Slivkins(2011)}]{pmlr-v19-slivkins11a}
Slivkins, A. 2011.
\newblock Contextual Bandits with Similarity Information.
\newblock In Kakade, S.~M.; and von Luxburg, U., eds., \emph{Proceedings of the 24th Annual Conference on Learning Theory}, volume~19 of \emph{Proceedings of Machine Learning Research}, 679--702. Budapest, Hungary: PMLR.

\bibitem[{Sun et~al.(2023)Sun, Xu, Tang, Wang, Lin, Gong, Shum, and Guo}]{thinkongraph}
Sun, J.; Xu, C.; Tang, L.; Wang, S.; Lin, C.; Gong, Y.; Shum, H.-Y.; and Guo, J. 2023.
\newblock Think-on-Graph: Deep and Responsible Reasoning of Large Language Model with Knowledge Graph.
\newblock arXiv:2307.07697.

\bibitem[{Sun et~al.(2024{\natexlab{a}})Sun, Qin, Li, Shen, Qiao, and Zhong}]{sun2024linear}
Sun, W.; Qin, Z.; Li, D.; Shen, X.; Qiao, Y.; and Zhong, Y. 2024{\natexlab{a}}.
\newblock Linear Attention Sequence Parallelism.
\newblock \emph{arXiv preprint arXiv:2404.02882}.

\bibitem[{Sun et~al.(2024{\natexlab{b}})Sun, Qin, Sun, Li, Li, Shen, Qiao, and Zhong}]{sun2024co2}
Sun, W.; Qin, Z.; Sun, W.; Li, S.; Li, D.; Shen, X.; Qiao, Y.; and Zhong, Y. 2024{\natexlab{b}}.
\newblock CO2: Efficient distributed training with full communication-computation overlap.
\newblock \emph{arXiv preprint arXiv:2401.16265}.

\bibitem[{Talmor and Berant(2018)}]{talmor2018web}
Talmor, A.; and Berant, J. 2018.
\newblock The web as a knowledge-base for answering complex questions.
\newblock \emph{arXiv preprint arXiv:1803.06643}.

\bibitem[{Tekin and Tur{\u{g}}ay(2018)}]{dominantMOMAB}
Tekin, C.; and Tur{\u{g}}ay, E. 2018.
\newblock Multi-objective contextual multi-armed bandit with a dominant objective.
\newblock \emph{IEEE Transactions on Signal Processing}, 66(14): 3799--3813.

\bibitem[{Touvron et~al.(2023)Touvron, Martin, Stone, Albert, Almahairi, Babaei, Bashlykov, Batra, Bhargava, Bhosale et~al.}]{touvron2023llama}
Touvron, H.; Martin, L.; Stone, K.; Albert, P.; Almahairi, A.; Babaei, Y.; Bashlykov, N.; Batra, S.; Bhargava, P.; Bhosale, S.; et~al. 2023.
\newblock Llama 2: Open foundation and fine-tuned chat models.
\newblock \emph{arXiv preprint arXiv:2307.09288}.

\bibitem[{Wanigasekara et~al.(2019)Wanigasekara, Liang, Goh, Liu, Williams, and Rosenblum}]{wanigasekara2019learning}
Wanigasekara, N.; Liang, Y.; Goh, S.~T.; Liu, Y.; Williams, J.~J.; and Rosenblum, D.~S. 2019.
\newblock Learning Multi-Objective Rewards and User Utility Function in Contextual Bandits for Personalized Ranking.
\newblock In \emph{IJCAI}, volume~19, 3835--3841.

\bibitem[{Weymark(1981)}]{GGI}
Weymark, J.~A. 1981.
\newblock Generalized Gini inequality indices.
\newblock \emph{Mathematical Social Sciences}, 1(4): 409--430.

\bibitem[{Xie et~al.(2022)Xie, Wu, Shi, Zhong, Scholak, Yasunaga, Wu, Zhong, Yin, Wang et~al.}]{xie2022unifiedskg}
Xie, T.; Wu, C.~H.; Shi, P.; Zhong, R.; Scholak, T.; Yasunaga, M.; Wu, C.-S.; Zhong, M.; Yin, P.; Wang, S.~I.; et~al. 2022.
\newblock UnifiedSKG: Unifying and Multi-Tasking Structured Knowledge Grounding with Text-to-Text Language Models.
\newblock In \emph{Proceedings of the 2022 Conference on Empirical Methods in Natural Language Processing}, 602--631.

\bibitem[{Xu et~al.(2024)Xu, Cruz, Guevara, Wang, Deshpande, Wang, and Li}]{KAPING}
Xu, Z.; Cruz, M.~J.; Guevara, M.; Wang, T.; Deshpande, M.; Wang, X.; and Li, Z. 2024.
\newblock Retrieval-Augmented Generation with Knowledge Graphs for Customer Service Question Answering.
\newblock \emph{arXiv preprint arXiv:2404.17723}.

\bibitem[{Yih et~al.(2016)Yih, Richardson, Meek, Chang, and Suh}]{yih2016value}
Yih, W.-t.; Richardson, M.; Meek, C.; Chang, M.-W.; and Suh, J. 2016.
\newblock The value of semantic parse labeling for knowledge base question answering.
\newblock In \emph{Proceedings of the 54th Annual Meeting of the Association for Computational Linguistics (Volume 2: Short Papers)}, 201--206.

\bibitem[{Yu et~al.(2022)Yu, Zhang, Ng, Zhu, Li, Wang, Hu, Wang, Wang, and Xiang}]{decaf}
Yu, D.; Zhang, S.; Ng, P.; Zhu, H.; Li, A.~H.; Wang, J.; Hu, Y.; Wang, W.; Wang, Z.; and Xiang, B. 2022.
\newblock Decaf: Joint decoding of answers and logical forms for question answering over knowledge bases.
\newblock \emph{arXiv preprint arXiv:2210.00063}.

\bibitem[{Zhang et~al.(2023)Zhang, Xiao, Liu, Dou, and Nie}]{bge}
Zhang, P.; Xiao, S.; Liu, Z.; Dou, Z.; and Nie, J.-Y. 2023.
\newblock Retrieve anything to augment large language models.
\newblock \emph{arXiv preprint arXiv:2310.07554}.

\bibitem[{Zhou, Li, and Gu(2020)}]{zhou2020neural}
Zhou, D.; Li, L.; and Gu, Q. 2020.
\newblock Neural contextual bandits with ucb-based exploration.
\newblock In \emph{International Conference on Machine Learning}, 11492--11502. PMLR.

\end{thebibliography}

\section{Appendix}

\subsection{Generalized Gini Index}
\label{GGI-detail}
To further leverage the unique strengths of each method and enhance the overall efficacy of the RAG pipeline. This necessitates the formulation of a multi-objective optimization problem.

%The Generalized Gini Index (GGI) provides a sophisticated measure for balancing various criteria in multi-objective optimization problems. 

The Generalized Gini Index (GGI) emerges as a crucial tool in this context, offering a sophisticated framework for equitably balancing diverse criteria in multi-objective optimization scenarios. The need for such an optimization arises from the varied and often conflicting objectives associated with different retrieval methods. For instance, while one method may excel in accuracy, another might offer benefits in terms of speed. The challenge, therefore, lies in achieving an optimal balance that maximizes the overall performance of the RAG system.

GGI has been characterized by \cite{GGI}. It encodes both efficiency as it is monotone with Pareto dominance and fairness as it is non-increasing with Pigou-Dalton transfers\cite{jenkins2017measurement}. Informally, a Pigou-Dalton transfer involves increasing a lower-valued objective while decreasing a higher-valued objective by an equivalent amount, without altering the order between the two objectives. This operation seeks to equilibrate the cost vector. Formally, the GGI adheres to the following fairness principle: for any \(x_i < x_j\),

\begin{equation}
\forall \epsilon \in (0, x_j - x_i), \quad G_w (\mathbf{x} + \epsilon \mathbf{e}_i - \epsilon \mathbf{e}_j) \leq G_w (\mathbf{x})
\end{equation}
where \(\mathbf{e}_i\) and \(\mathbf{e}_j\) are two vectors of the canonical basis. Consequently, among vectors with an equal sum, the optimal cost vector (with respect to GGI) is the one that features uniform values across all objectives, provided that such a distribution is feasible.

\subsection{MAB enhanced RAG systems with LLM variants}
\label{variants}

\begin{figure}[ht!]
 \includegraphics[width=\linewidth]{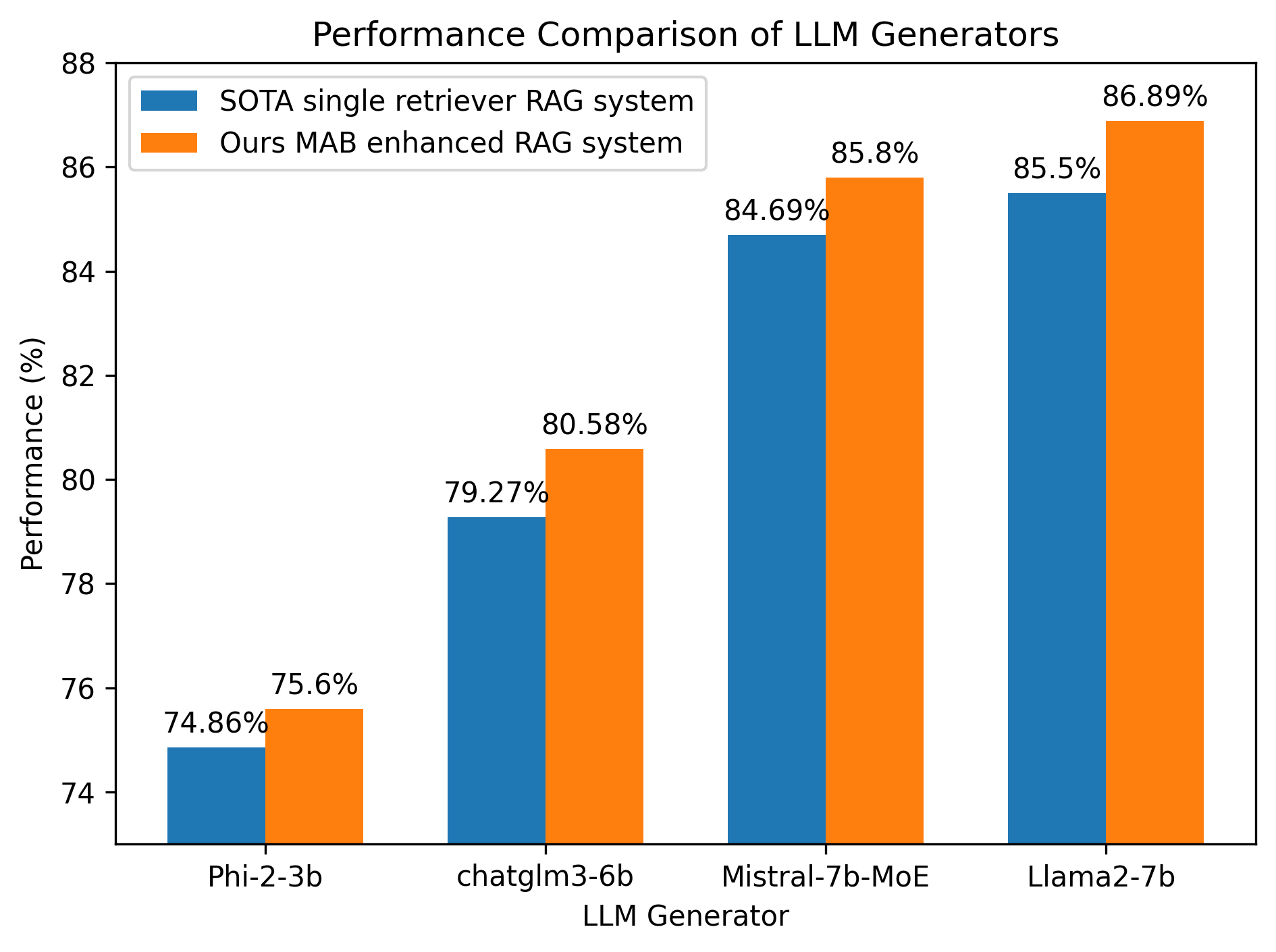}
 \caption{MAB enhanced RAG systems with LLM variants under stationary environments}
 \label{fig:llm_variant}
\end{figure}

We evaluated our MAB-enhanced RAG system using various LLM generators, including Chatglm3 \cite{du2021glm} and Mistral \cite{jiang2023mistral}. The results, depicted in \cref{fig:llm_variant}, demonstrate that our system significantly improves RAG performance compared to traditional systems that utilize only a single retriever. This enhancement underscores the robustness and adaptability of our MAB-enhanced RAG approach.

\subsection{Implementation detail}
\label{app:imp}
To leverage the strengths of various retrieval methods, our MAB-enhanced RAG system, along with other router-based RAG approaches, utilizes DECAF\cite{decaf}, ChatKBQA\cite{chatkbqa}, and Reason-on-Graph\cite{Rog} as its action space.
In \cref{tab:station_comp}, the term ``Ensemble'' and in \cref{tab:non-station}, ``Retrieval Ensemble'' refer to configurations where multiple retrievers operate in a complementary manner.
 Following the methodology outlined in \cite{llamaIndexDefineSelector} the LLM Router is referring to a prompt-based GPT-4 (32k) router accessed via the OpenAI API.
Approaches such as those proposed by \cite{sun2024co2,sun2024linear} can be utilized to accelerate the training process.

\subsection{Case study}

    % To further illustrate the effectiveness of our system, we present a case study where the user queries, "What are some books that Mark Twain wrote?" This question is challenging in terms of achieving high recall since all retrieval methods can provide related context, but not all can accurately list the books.
    % Our MAB-enhanced RAG system effectively selects the appropriate methods to achieve the highest recall, significantly outperforming individual retrieval approaches.

\begin{figure}[ht!]
         \includegraphics[width=\linewidth]{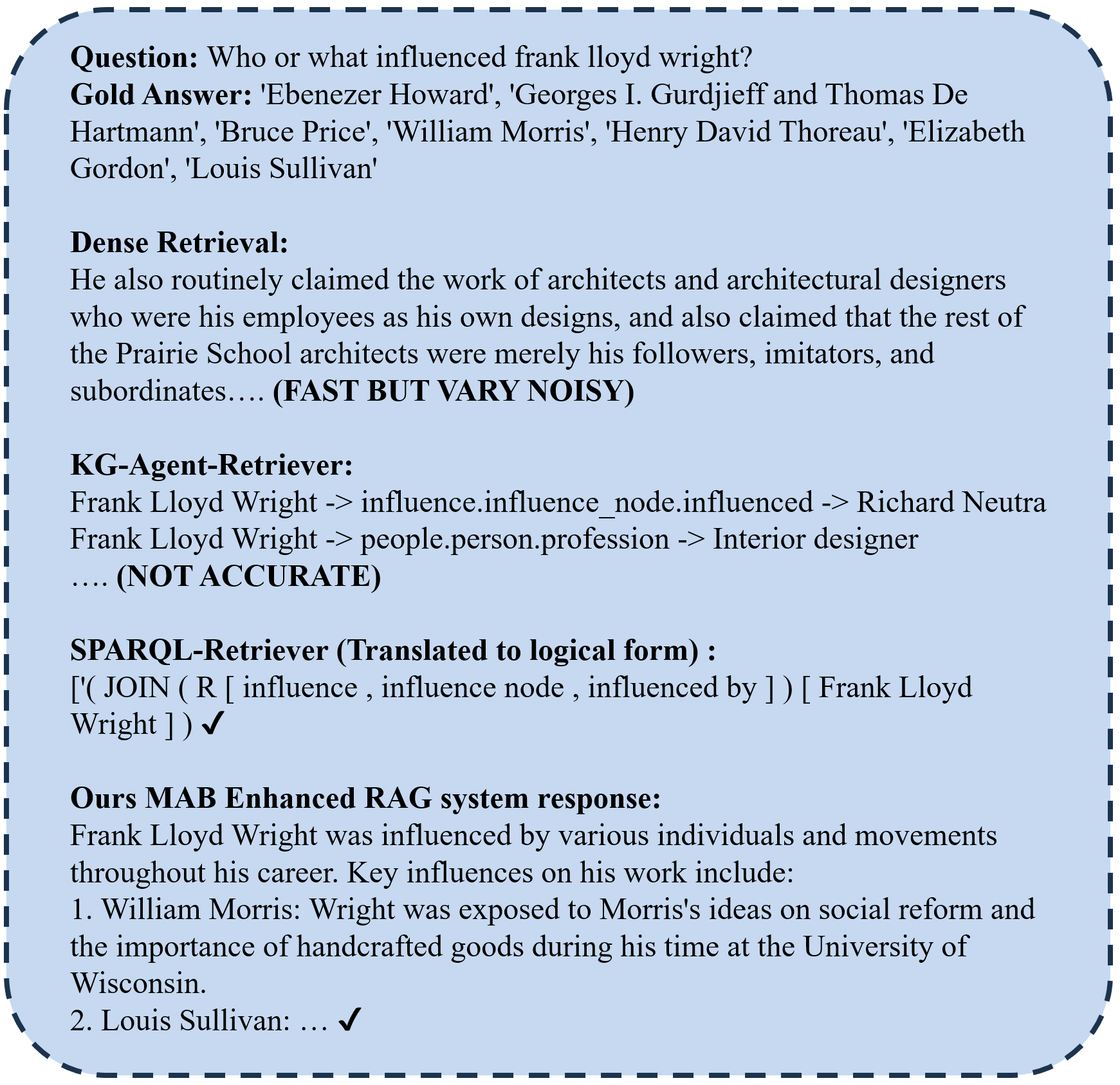}
         \vspace{-2mm}
         \caption{SPARQL-Based Retriever gives the most accurate context}
         \label{fig:example_webqsp}
    \end{figure}

    We further present a case study \cref{fig:example_webqsp} to illustrate the effectiveness of our system, the query inquires about influences on Frank Lloyd Wright. While dense retrieval provides a fast response, it introduces noisy and irrelevant data. The LLM agent retriever, though accessing structured knowledge graphs, fails to deliver accurate information, focusing instead on peripheral details like professions. However, the SPARQL-retriever can give a very accurate query code for this question with clear conditions, and successfully fetch all results that support our systems to offer a nuanced and accurate response.

\subsection{Experiment of additional non-stationarity}

As shown in \cref{tab:half_column_table}, we simulate the scenario where one of the retrieval services fails, (unable to return any documents) and requires router to swiftly reorganize the queries to other retrieval methods.

\begin{table}[]
    \centering
    \resizebox{\columnwidth}{!}{%
        \begin{tabular}{@{}cccc@{}}
            \toprule
            \textbf{Method} & \textbf{Test Hit} & \textbf{Test Recall} & \textbf{Test Retrieval Delay} \\ 
            \midrule
            GGI-MO-MAB (offline) & 61.70 ± 2.08 & 42.50 ± 2.62 & 4.84 ± 0.81 \\
            Static NN-Router     & 64.99 ± 1.24 & 46.23 ± 1.58 & 5.44 ± 0.54 \\
            GGI-MO-MAB (online)  & 77.55 ± 2.40 & 60.07 ± 3.29 & 11.23 ± 3.21 \\
            \bottomrule
        \end{tabular}
    }
    \caption{Performance comparison under system degradation scenario}
    \label{tab:half_column_table}
\end{table}

\vspace{-2mm}
\subsection{Retrieval methods on KG}
\label{Retrieval-methods-on-KG}
\textbf{Dense retrieval}: Dense retrieval methods \cite{bge,DPR} standardize and segment diverse document formats like PDF, HTML, Word, and Markdown into plain text, which is then transformed into vector embeddings for efficient searching \cite{DPR}. Utilizing a pre-trained language model \cite{bert}, DPR creates dense embeddings from question-passage pairs, significantly enhancing accuracy over BM25 and ORQA in open Natural Questions. Additionally, recent developments by \cite{bge} have tailored embedding models to meet the varied retrieval demands of LLMs with techniques like knowledge distillation and multi-task fine-tuning.

There are two main approaches to applying dense retrieval to Knowledge Graph data: directly searching the textual data (e.g. Wikipedia) that constitutes the KG \cite{ennen2023hierarchical,DPR}, and linearizing KGs into text corpora \cite{decaf,xie2022unifiedskg}, which translates structured knowledge into natural language form.

Dense retrieval methods are typically fast with pre-built vector bases, but it rely on embedding models for query reasoning, often falling short in complex, multi-hop retrieval tasks that demand greater analytical depth \cite{behnamghader2023can}.

% 可以进行一定程度的推理
\textbf{KG Query Language Retrieval}: 
% say this method is more explainable
Due to the structured representation and storage, it is efficient to access structured data using query languages (e.g. SPARQL) or specific algorithms (e.g., triple search for knowledge graphs).
ChatKBQA \cite{chatkbqa} proposes generating the logical form with fine-tuned LLMs first, then retrieving and replacing entities and relations through an unsupervised retrieval method, which improves both generation and retrieval more straightforwardly.
StructGPT \cite{structgpt} constructs the specialized interfaces to collect relevant evidence from structured data, and let LLMs concentrate on the reasoning task based on the collected information \cite{chentraining}.

\textbf{KG agent-based retrieval}: Differing from SPARQL generator retrieval, LLM agent-based retrieval adopts a tightly coupled "LLM-KG" paradigm. In this setup, agents such as LLMs navigate through relations and entities on Knowledge Graphs and construct a reasoning path for answering queries. 
\cite{thinkongraph} introduces a new approach called Think-on-Graph (ToG), in which the LLM agent iteratively executes beam search on KG, discovers the most promising reasoning paths, and returns the most likely reasoning results.
\cite{Rog} presents a planning retrieval-reasoning framework, where RoG first generates relation paths grounded by KGs as faithful plans. These plans are then used to retrieve valid reasoning paths from the KGs for LLMs to conduct faithful reasoning\cite{Molecule}.

\subsection{RAG Prompts Used in Experiments}
\begin{lstlisting}[frame=single, caption=RAG prompt, numbers=none, linewidth=\linewidth, breaklines=true, breakatwhitespace=false]
System prompt:
Based on the context information provided, and not on prior knowledge, please answer the given question.
Context: {context}
Question:
Please answer the query: {query}
\end{lstlisting}

\subsection{Statistics of datasets}
\begin{table}[h!]
\centering
\caption{Statistics of datasets.}
\begin{tabular}{cccc}
\toprule
\textbf{Datasets} & \textbf{Train} & \textbf{Test} & \textbf{Max hop} \\ \midrule
WebQSP            & 2,826            & 1,628           & 2                  \\ \midrule
CWQ               & 27,639           & 3,531           & 4                  \\ \bottomrule
\end{tabular}
\label{tab:statistic}
\end{table}

% \section{Acknowledgments}

\end{document}